\documentclass[conference]{IEEEtran}
\IEEEoverridecommandlockouts
\usepackage{subfigure}
\usepackage{cite}
\usepackage{amsmath,amssymb,amsfonts}
\usepackage{algorithmic}
\usepackage{textcomp}
\usepackage{caption}
\usepackage{graphicx}
\usepackage{multirow}

\usepackage[table,xcdraw]{xcolor}

\def\BibTeX{{\rm B\kern-.05em{\sc i\kern-.025em b}\kern-.08em
    T\kern-.1667em\lower.7ex\hbox{E}\kern-.125emX}}
\usepackage{comment}
\usepackage{authblk}

\usepackage{amssymb}
\usepackage{fancyhdr}
\fancypagestyle{firstpage}
{
    \fancyhead[L]{\footnotesize © 2023 IEEE.  Personal use of this material is permitted.  Permission from IEEE must be obtained for all other uses, in any current or future media, including reprinting/republishing this material for advertising or promotional purposes, creating new collective works, for resale or redistribution to servers or lists, or reuse of any copyrighted component of this work in other works. This paper is accepted at the 26th International Symposium on Design and Diagnostics of Electronic Circuits and Systems (DDECS) 2023.}
    \fancyhead[R]{}
}
\begin{document}

\title{APPRAISER: DNN Fault Resilience Analysis Employing Approximation Errors\\ 
}
\author[1]{Mahdi Taheri}
\author[1]{Mohammad Hasan Ahmadilivani}
\author[1]{Maksim Jenihhin}
\author[1,2]{Masoud Daneshtalab}
\author[1]{Jaan Raik}
\affil[1]{Tallinn University of Technology, Tallinn, Estonia}
\affil[2]{Mälardalen University, Västerås, Sweden}
\affil[1]{mahdi.taheri@taltech.ee}

\maketitle
\thispagestyle{firstpage}

\begin{abstract}
Nowadays, the extensive exploitation of Deep Neural Networks (DNNs) in safety-critical applications raises new reliability concerns. In practice, methods for fault injection by emulation in hardware are efficient and widely used to study the resilience of DNN architectures for mitigating reliability issues already at the early design stages. However, the state-of-the-art methods for fault injection by emulation incur a spectrum of time-, design- and control-complexity problems. To overcome these issues, a novel resiliency assessment method called APPRAISER is proposed that 
applies functional approximation for a non-conventional purpose and employs approximate computing errors for its interest. 
By adopting this concept in the resiliency assessment domain, APPRAISER provides thousands of times speed-up in the assessment process, while keeping high accuracy of the analysis. In this paper, APPRAISER is validated by comparing it with state-of-the-art approaches for fault injection by emulation in FPGA. By this, the feasibility of the idea is demonstrated, and a new perspective in resiliency evaluation for DNNs is opened.
\end{abstract}


\vspace{-0.12cm}\section{Introduction}

\vspace{-0.1cm}
In recent years, Deep Neural Networks (DNNs) surpassed human-level precision\cite{silver2017mastering} that made them attractive for several safety-critical applications such as autonomous driving \cite{al2017deep, taheri2022dnn}.

Faults that can be caused by soft errors, aging, etc., are the source of threatening the reliability of DNN inference hardware accelerators. Here, \emph{soft errors}, are of particular concern for researchers in the industry and academia. It is a class of faults caused by ionized particles hitting transistors that can flip a logic value in a memory cell or a logic gate. 

In today's applications, network parameters, e.g., weights, occupy most of the inference accelerator's areal footprint, making them natural targets for soft-errors-caused disturbances. Unlike other logic structures, DNNs are known to be relatively resilient to transient faults. However, in practice, such faults still may cause a significant accuracy drop in DNNs because of the large area and memory requirements for the state-of-the-art DNNs accelerators.
Although numerous techniques have been proposed recently to evaluate the architectural fault resilience of DNNs, they are still rather costly. 
Throughout the literature, Fault Injection (FI) is the most commonly used method for resilience evaluation of DNNs.

Fault injection by emulation in hardware, usually in FPGAs, is widely adopted by the industry \cite{ibrahim2020soft} because of its ability to evaluate real-scale DNN accelerator designs with significantly shorter run times compared to software-based simulation.

\begin{figure}
    \centering
    \includegraphics[width = 0.35\textwidth]{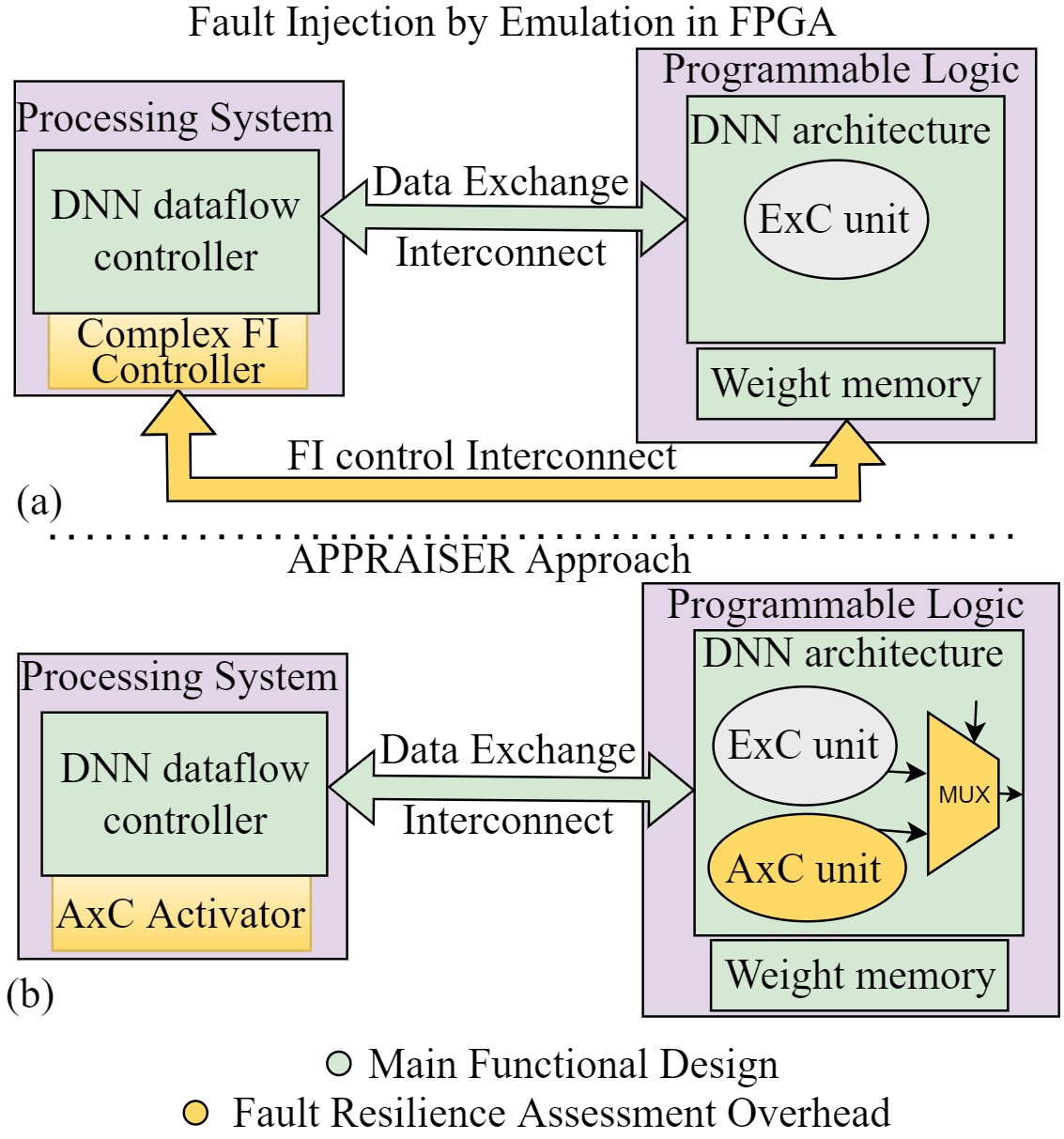}
    \caption{DNN fault resiliency assessment methods: (a) Fault injection by emulation in FPGA; (b) APPRAISER approach using errors by AxC units.}
    \label{fiflow}\label{ARAM:methodology}
\end{figure}

However, the state-of-the-art approaches for fault injection by emulation in hardware imply iterative procedures for each injected fault, including numerous extra memory accesses, which make them time-consuming and imply complex implementation. Fig. \ref{fiflow}(a) illustrates the execution overheads of the  general flow of FI by emulation in hardware. In particular, such an iterative approach is breaking the pipeline and requires a complex FI Controller and an extra FI control interconnect \cite{hsueh1997fault, khoshavi2020compression, khoshavi2020shieldenn, fiji}. 
Fig. \ref{fiflow} (b) illustrates the proposed approach APPRAISER, which allows reducing the fault resiliency assessment overheads.  
The ability to tolerate the impact of faults on the output accuracy is called \textit{fault resiliency} and, in practice, it is one of the contributors to the final DNN accelerators' reliability  \cite{taheri23}.

In this paper, our contribution is a novel method of fault resiliency analysis for DNN architectures that applies functional approximation for a non-conventional purpose and harnesses approximate computing errors for its interest. To the best of our knowledge, for the first time, \textit{Approximate Computing} (AxC) units are adopted to improve the processing time-, design-, and control-complexity for DNN fault resiliency analysis process.

APPRAISER provides a rapid exploration of different options of the network architecture, training, dataset, etc., to study the fault resilience of the DNNs. In particular, it enables efficient analysis of subsequent layers' resilience to faults in the weights of a compromised layer. 

The new method has the following advantages:
\begin{itemize}

    \item It eliminates the need for designing and deploying an extra complex controller for the fault injection procedure. A simple approximate units enabling circuitry (AxC Activator) is employed instead. 

    \item The inference pipeline process executes a batch of inputs with no need to break this process. 

    \item The resilience assessment process is performed without an extra interconnect for weight sampling.
    \item The proposed approach is not iterative for each potential fault location, unlike the traditional fault injection. Thus, the analysis complexity is vastly reduced.

    \end{itemize}

The rest of the paper is organized as follows. An analysis of Related Works in Section II is followed by the new methodology presented in Section III. The experimental results, along with their discussion, are presented in section IV. Finally, this work is concluded in Section V.

\vspace{-0.1cm}\section{Related works}

\vspace{-0.1cm}

The extensive growth of the memory footprint size in today's practical DNN inference HW accelerators increases the chances of soft errors' occurrences causing prediction failures. 
Even a minor change in the DNN architecture may cause a notable difference in the DNNs' architectural fault resiliency \cite{taheri23}.
Evaluating the resiliency of DNNs with FI by emulation in hardware is a practical method used today by the industry. There are several works emulating fault injection on FPGAs as a hardware platform. 

Fiji-FIN \cite{fiji} is one of such DNNs' resiliency evaluation frameworks. 
It considers the model's accuracy degradation as a metric to study the impact of soft errors on the network's parameters, such as weights and activation. 
Unfortunately, it implies severe effort for designing the fault injection campaigns. For each single fault injection, the execution of the inference should be halted for manipulating the DNN parameters, and it has to be resumed thereafter. It means that the classification time for a batch of inputs should be interrupted to apply fault injection between the classification process of two consecutive inputs.

A similar method is also used in \cite{khoshavi2020compression, khoshavi2020shieldenn}. These works also propose injecting transient faults into on-chip memories of the design implemented on the FPGA. In these works, the bit stream file of the FPGA is obtained by a High-Level Synthesis (HLS) tool and imported to the FPGA. While the system is running, the faults are generated and injected by the embedded processor and the reliability is evaluated in comparison with the golden model.

In contrast to the works mentioned above, this paper proposes a novel non-iterative fault resilience analysis by exploiting the approximation errors instead of fault injection

It enables keeping the inference pipeline process to be executed on a batch of inputs unbroken.

\vspace{-0.1cm}\section{Proposed Method APPRAISER}
The proposed approach for applying errors of approximate computing units for DNN fault resiliency assessment is outlined in Fig. \ref{ARAM:methodology}(b). An AxC Activator unit on the Processing System (PS) side enables the AxC units to induce errors. These units are AxC multipliers in the multiply-and-accumulate units (MACs), in the targeted (mimicked to be compromised) layer of the DNN. This activator controls the multiplexers on the Programmable Logic (PL) side to switch between the exact implementation of the units (for the functional mode) and the approximated one (for the resiliency assessment mode). Then, the user runs the inference just once for the validation dataset and stores the results of the layers' outputs. 

\begin{figure}[h]
    \centering
    \includegraphics[width = 0.5\textwidth]{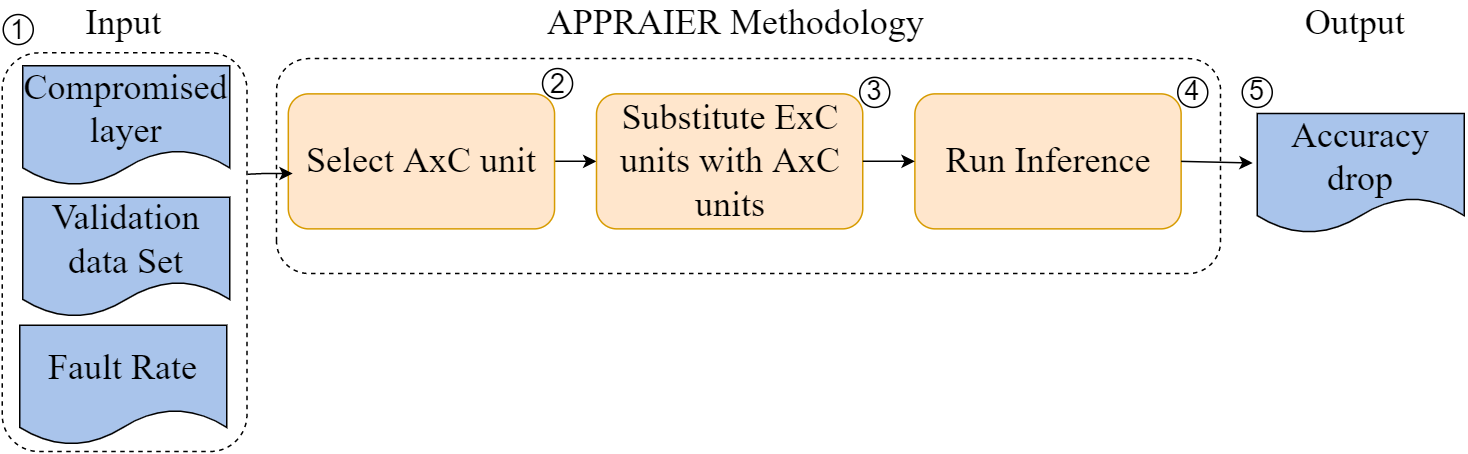}
    \caption{APPRAISER Methodology}
    \label{method}
\end{figure}

The flow of APPRAISER method is depicted in Fig. \ref{method}. 

Step 1 is the initialization that includes the selection of the compromised layer (e.g. one by one in the DNN structure), the validation testset (i.e. DNN inputs), and the assumed application-specific fault rate. In Step 2, suitable AxC units are selected. For example, in this work, we used AxC multipliers from the EvoApproxLib library \cite{evoapprox16}. 
 Further, a set of ExC units are substituted with the AxC units in the network architecture (Step 3). The DNN inference is executed keeping the pipeline of the network and the \emph{DNN output accuracy drop} is reported. It is used as the main DNN fault resilience analysis metric. The more the accuracy drops with the induced errors, the less fault-resilient the given DNN implementation is.

\begin{figure}[h]
    \centering
    \includegraphics[width = 0.55\linewidth]{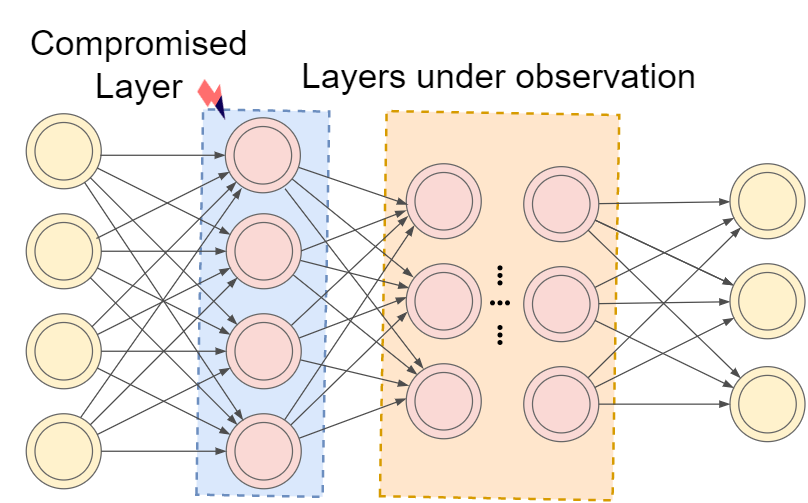}
    \caption{APPRAISER assessment flow: Compromised layer in the presence of faults in weights vs. layers under resiliency test}
    \label{ARAM}
\end{figure}

In a traditional application of AxC, the approximation of hardware components is based on their inexact implementation that creates a functionally tolerable mismatch with the specification while providing gains in compute-efficiency. In practice, there is an error induced by approximation that can also be employed to mimic the error caused by a fault in the inputs of a logic circuit that is propagated to the output. Such approximation-induced errors affect their corresponding outputs, which are also connected to several other neurons in subsequent layers (as their activation) (Fig. \ref{ARAM}). 

The characteristics of the approximation-induced errors can be assessed by several metrics, normalized error, number of flipped bits, and their impact on the neural network classification accuracy drop. In this study, we rely on the following simple set of metrics:
\begin{enumerate}
    \item \textit{normalized error}: calculated as the average error on the output of each layer by subtracting the neurons' outputs of that layer from the golden output and dividing all the error values to the maximum value;
    \item \textit{network accuracy and recall drop}: calculated by executing the network under different circumstances (faulty vs approximated) over the test set;
    \item \textit{bitflips in subsequent layers}: calculated by comparing all bits in the next layers' outputs with the golden model and counting the bits that do not match as flipped bits.

    \end{enumerate}

The main objective of APPRAISER is the study of the resiliency of DNN architecture layers to faults that might occur in the weights of a compromised layer. 
By using this method, the user can rapidly explore the options of network architecture, training, dataset, etc., in terms of fault resiliency analysis.

Unlike some other frameworks (e.g., FijiFin \cite{fiji}, APPRAISER does not support assessing the reliability of the network to faults in the activations and DNN neurons and currently is only aimed at resiliency to faults in stored DNN weights. 
Other limitations are a lower diagnostic capability and implicit correspondence to traditional fault injection based metrics (e.g. in standards). 

\vspace{-0.12cm}\section{Experimental Results}
\subsection{Evaluation Methodology}
\begin{figure}
    \centering
    \includegraphics[width = 0.4\textwidth]{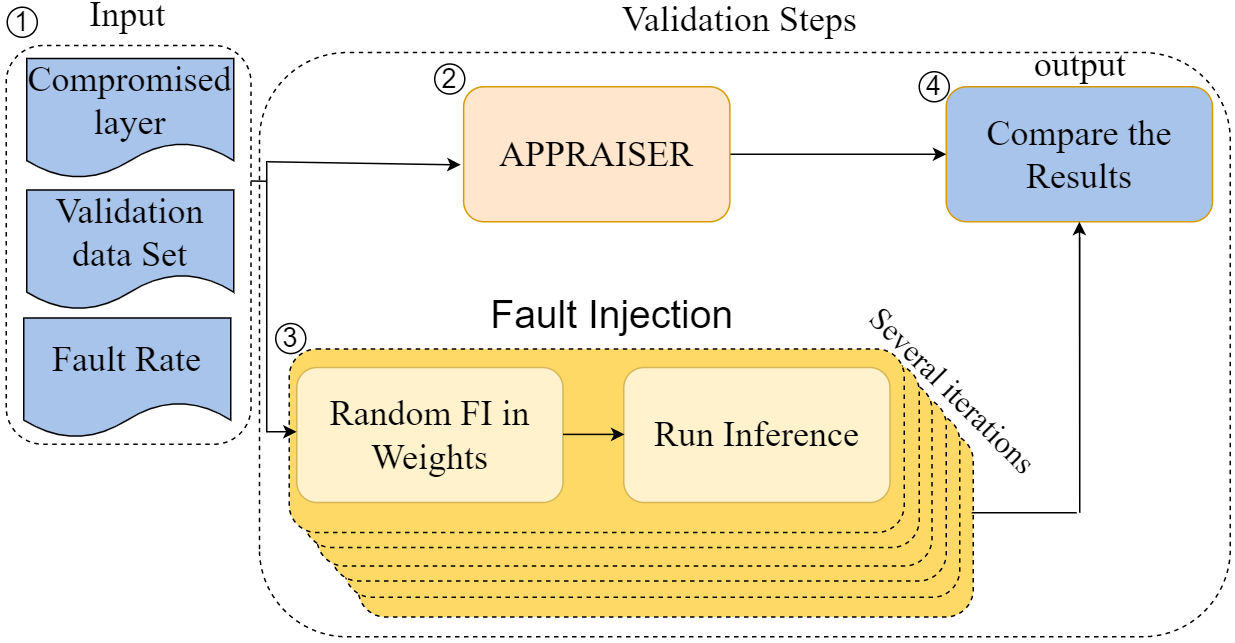}
    \caption{APPRAISER evaluation flow}
    \label{experiment}
\end{figure}

The flow to evaluate the proposed method is illustrated in Fig. \ref{experiment}. Here, Steps 1 and 2 repeat the APPRAISER method execution (Fig. \ref{method}).  

The list of candidate approximate multipliers from the EvoApproxLib library \cite{evoapprox16} was narrowed down with several relevant metrics 
adopted from EvoApproxLib with the main focus on two established features (Variance of Error Distance (Var-ED) and Root Mean Square (RMS-ED)) presented in \cite{ansari2019improving}. These two metrics are the most critical approximation-induced errors' features for the performance of an AxC unit in DNNs. Based on these metrics, Mult8s\_1KX2 (further referred to as \emph{Mult1}) and Mult8s\_1KRC (\emph{Mult2}) multipliers are selected for the experiment.

For the reference part, the fault resiliency evaluation is repeated on the original network instrumented for a state-of-the-art FI method \cite{fiji} (Step 4). Two fault models are considered in this study:
\begin{itemize}
    \item Injection of a \emph{single bitflip} at a random location in all weight bits of the compromised layer for every input in the DNN validation test set,
    \item Injection of \emph{double bitflips} in weights of the compromised layer for every input in the DNN validation test set.
\end{itemize}

For each fault model, the experiment is repeated for 1000 random faults per image are considered to reach the 95\% FI confidence level according to the statistical fault injection approach \cite{leveugle2009statistical} and in the end, the average accuracy of all repetitions is reported. Finally, the DNN accuracy drops as a result of applying approximation and fault injection along with normalized error, and the number of flipped bits are compared (Step 5).

\begin{figure*}[h]
     \centering
     \begin{minipage}[b]{0.45\textwidth}
         \centering
         \includegraphics[width=1\textwidth]{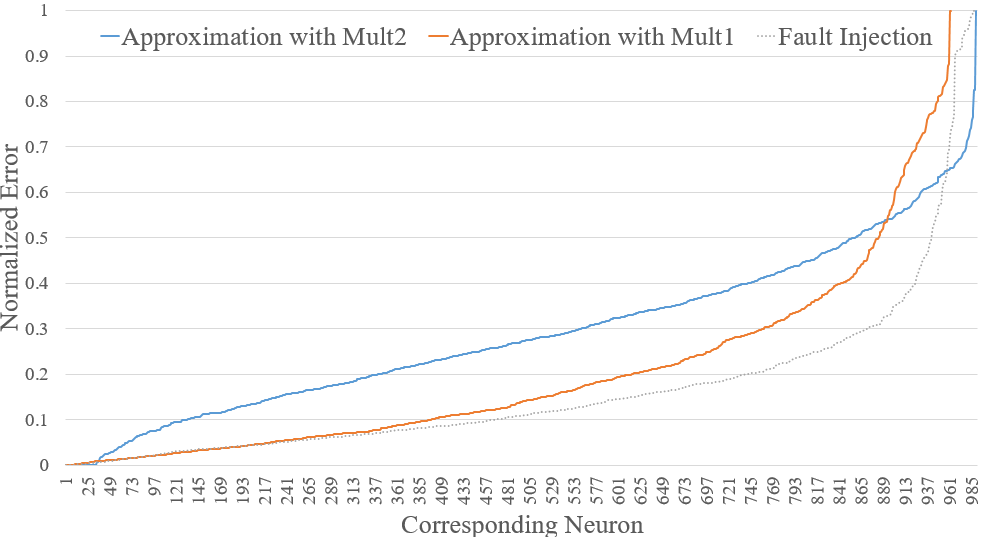}
         \caption{Normalized output error of Conv2: Applying AxC and fault injection on the Conv1}
         \label{res1}
     \end{minipage}
     \hfill
     \begin{minipage}[b]{0.45\textwidth}
         \centering
         \includegraphics[width=1\textwidth]{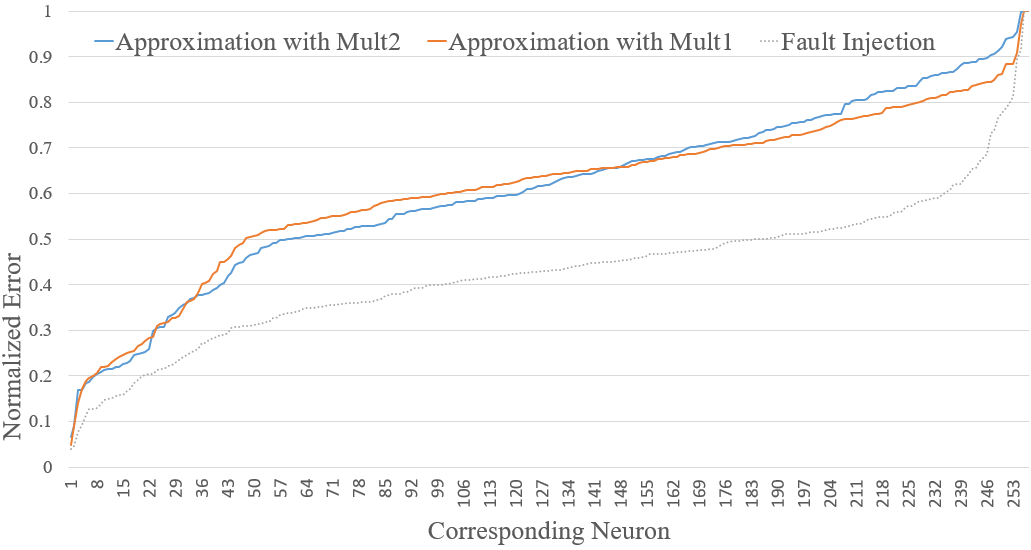}
         \caption{Normalized output error of Pool2: Applying AxC and fault injection on the Conv1}
         \label{res2}
     \end{minipage}
\end{figure*}

\vspace{-0.1cm}\subsection{Experimental Setup}
To evaluate the feasibility of the proposed method, a simple Convolutional Neural Network (CNN) with two convolutional layers, two max-pooling, and one Fully-Connected (FC) layer was implemented and trained. The simulations were performed on an Intel® Core™ i7-6800K CPU @ 3.40GHz × 12, and the proposed method was implemented with Python 3. The hardware synthesis and implementation results are produced by the Xilinx Vivado HLS tool on a Xilinx Spartan-7 FPGA (xc7s100-fgga676-1) at 100 MHz operational frequency.

The CNN under study is trained on a dataset of 2000 images of animals (cats and dogs) and humans for binary classification. The accuracy of the network over the test set (including 450 images of animals and humans) is 93.34\%. Bit truncation quantization is applied in network parameters during training and data precision is reduced to 8-bit. 

\begin{table*}[h]
\caption{Bitflips and Accuracy/Recall drop induced by APPRAISER vs the reference fault injection method}
\label{tab:my-table} \label{bit_flip_conv1}
\resizebox{1\textwidth}{!}{
\tiny
\centering
\begin{tabular}{|c|cccccc|}
\hline
\multirow{3}{*}{Affected/Measured Layers} & \multicolumn{6}{c|}{Bitflips in subsequent layers}                                                                                                                                                                                                                                                                                                                                                                                                                                                                                                                                                         \\ \cline{2-7} 
                                          & \multicolumn{3}{c|}{Injection of a single fault}                                                                                                                                                                                                                                                               & \multicolumn{3}{c|}{Injection of a double fault}                                                                                                                                                                                                                                          \\ \cline{2-7} 
                                          & \multicolumn{1}{c|}{\begin{tabular}[c]{@{}c@{}}Fault Injection\\ (reference)   {[}\%{]}\end{tabular}} & \multicolumn{1}{c|}{\begin{tabular}[c]{@{}c@{}}Approximation with \\ MULT1 {[}\%{]}\end{tabular}} & \multicolumn{1}{c|}{\begin{tabular}[c]{@{}c@{}}Approximation with\\  MULT2 {[}\%{]}\end{tabular}} & \multicolumn{1}{c|}{\begin{tabular}[c]{@{}c@{}}Fault Injection\\ (reference){[}\%{]}\end{tabular}} & \multicolumn{1}{c|}{\begin{tabular}[c]{@{}c@{}}Approximation with \\ MULT1 {[}\%{]}\end{tabular}} & \begin{tabular}[c]{@{}c@{}}Approximation with\\  MULT2 {[}\%{]}\end{tabular} \\ \hline
Conv1/Conv1                               & \multicolumn{1}{c|}{10.00}                                                                             & \multicolumn{1}{c|}{9.97}                                                                         & \multicolumn{1}{c|}{9.98}                                                                         & \multicolumn{1}{c|}{9.99}                                                                              & \multicolumn{1}{c|}{10.00}                                                                        & 9.99                                                                         \\ \hline
Conv1/Pool1                               & \multicolumn{1}{c|}{9.03}                                                                              & \multicolumn{1}{c|}{9.03}                                                                         & \multicolumn{1}{c|}{9.03}                                                                         & \multicolumn{1}{c|}{9.06}                                                                              & \multicolumn{1}{c|}{9.06}                                                                         & 9.05                                                                         \\ \hline
Conv1/Conv2                               & \multicolumn{1}{c|}{16.73}                                                                             & \multicolumn{1}{c|}{16.72}                                                                        & \multicolumn{1}{c|}{16.74}                                                                        & \multicolumn{1}{c|}{16.74}                                                                             & \multicolumn{1}{c|}{16.74}                                                                        & 16.74                                                                        \\ \hline
Conv1/Pool2                               & \multicolumn{1}{c|}{16.40}                                                                             & \multicolumn{1}{c|}{16.45}                                                                        & \multicolumn{1}{c|}{06.50}                                                                        & \multicolumn{1}{c|}{16.55}                                                                             & \multicolumn{1}{c|}{16.50}                                                                        & 16.45                                                                        \\ \hline
Conv1/FC                                  & \multicolumn{1}{c|}{9.25}                                                                              & \multicolumn{1}{c|}{9.25}                                                                         & \multicolumn{1}{c|}{8.50}                                                                         & \multicolumn{1}{c|}{9.30}                                                                              & \multicolumn{1}{c|}{9.30}                                                                         & 9.30                                                                         \\ \hline
Conv2/Conv2                               & \multicolumn{1}{c|}{16.71}                                                                             & \multicolumn{1}{c|}{16.72}                                                                        & \multicolumn{1}{c|}{16.71}                                                                        & \multicolumn{1}{c|}{16.76}                                                                             & \multicolumn{1}{c|}{16.74}                                                                        & 16.74                                                                        \\ \hline
Conv2/Pool2                               & \multicolumn{1}{c|}{16.40}                                                                             & \multicolumn{1}{c|}{16.45}                                                                        & \multicolumn{1}{c|}{16.41}                                                                        & \multicolumn{1}{c|}{16.50}                                                                             & \multicolumn{1}{c|}{16.50}                                                                        & 16.50                                                                        \\ \hline
Conv2/FC                                  & \multicolumn{1}{c|}{10.10}                                                                             & \multicolumn{1}{c|}{8.50}                                                                         & \multicolumn{1}{c|}{7.80}                                                                         & \multicolumn{1}{c|}{10.10}                                                                             & \multicolumn{1}{c|}{9.30}                                                                         & 8.30                                                                         \\ \hline
Affected Layer                            & \multicolumn{6}{c|}{DNN Accuracy/Recall drop}                                                                                                                                                                                                                                                                                                                                                                                                                                                                                                                                                              \\ \hline
Conv1                                     & \multicolumn{1}{c|}{2.3/4.7}                                                                           & \multicolumn{1}{c|}{2.7/8.0}                                                                      & \multicolumn{1}{c|}{2.2/6.7}                                                                      & \multicolumn{1}{c|}{4.7/14.0}                                                                          & \multicolumn{1}{c|}{5.8/17.4}                                                                     & 4.2/12.7                                                                     \\ \hline
Conv2                                     & \multicolumn{1}{c|}{1.8/6.0}                                                                           & \multicolumn{1}{c|}{1.6/5.0}                                                                      & \multicolumn{1}{c|}{2.7/8.0}                                                                      & \multicolumn{1}{c|}{9.1/26.4}                                                                          & \multicolumn{1}{c|}{9.1/26.4}                                                                     & 8.9/26.7                                                                     \\ \hline
\end{tabular}}

\end{table*}

\begin{table}[h!]
\caption{Overheads of APPRAISER vs the reference fault injection method (Conv1 layer)}
\label{hard}
\resizebox{.5\textwidth}{!}{
\centering
\begin{tabular}{|ccccc|}
\hline
\multicolumn{1}{|c|}{Network}                                                                    & \multicolumn{1}{c|}{\begin{tabular}[c]{@{}c@{}}Area LUT\\ utilization\end{tabular}} & \multicolumn{1}{c|}{\begin{tabular}[c]{@{}c@{}}Analysis Control\\ Circuitry\end{tabular}}     & \multicolumn{1}{c|}{Interconnects}                                                                  & \begin{tabular}[c]{@{}c@{}}DNN execution\\ time in FPGA\end{tabular} \\ \hline
\multicolumn{1}{|c|}{Base CNN}                                                                   & \multicolumn{1}{c|}{12\%}                                                           & \multicolumn{1}{c|}{N/A}                                                                      & \multicolumn{1}{c|}{\begin{tabular}[c]{@{}c@{}}Data Exchange \\ Interconnect\end{tabular}}          & 131ms                                                            \\ \hline
\multicolumn{5}{|c|}{Fault Resilience Assessment}                                                                                                                                                                                                                                                                                                                                                                                                                   \\ \hline
\multicolumn{1}{|c|}{\begin{tabular}[c]{@{}c@{}}CNN instrumented \\ with FI\end{tabular}}        & \multicolumn{1}{c|}{\textbf{23\%}}                                                  & \multicolumn{1}{c|}{Complex FI Controller}                                                    & \multicolumn{1}{c|}{\begin{tabular}[c]{@{}c@{}}(Data Exchange + FI) \\ Interconnect\end{tabular}}   & 632,000ms                                                            \\ \hline
\multicolumn{1}{|c|}{\begin{tabular}[c]{@{}c@{}}CNN instrumented\\  with APPRAISER\end{tabular}} & \multicolumn{1}{c|}{$\sim$29\%}                                                     & \multicolumn{1}{c|}{\textbf{\begin{tabular}[c]{@{}c@{}}Simple AxC \\ Activator\end{tabular}}} & \multicolumn{1}{c|}{\textbf{\begin{tabular}[c]{@{}c@{}}Data Exchange \\ Interconnect\end{tabular}}} & \textbf{131ms}                                                   \\ \hline
\end{tabular}}
\end{table}

\vspace{-0.12cm}\subsection{Evaluation Results}
The similarity of the fault resiliency analysis results by fault injection emulation and using the APPRAISER method is analyzed using the metrics identified in Section III. 

Fig. \ref{res1} illustrates \textit{normalized error} distribution in the output of the second convolutional layer (Conv2), in the presence of random double faults in the first convolution layer (dashed grey) vs errors induced by approximate multipliers (Mult1 solid orange, Mult2 solid blue) enabled in the first convolution layer respectively. Fig. \ref{res2} reports the result of applying FI and APPRAISER on the same convolutional layer and its impact on the second pooling layer of the network. These results demonstrate the similarity of the trends in error propagation by the proposed and the reference methods.

Table \ref{tab:my-table} reports fault resiliency assessment by the proposed and the reference methods using the \textit{bitflips in subsequent layers} and the \textit{DNN accuracy and recall drop} metrics. These results also demonstrate the strong similarity of the trends in error propagation by the proposed and the reference methods.

Table \ref{hard} demonstrates that although APPRAISER is more resource hungry, it is vastly faster than the reference fault injection by emulation method. It should be noted that the extra resources required by APPRAISER or FI are used only for the fault resiliency analysis phase and cleaned out from the final inference accelerator. In this example, the original CNN occupies 12\% of the FPGA resources (LUTs). The CNN instrumented with APPRAISER occupies 29\% of the FPGA resources and provides the accuracy/recall drop measurement for fault resiliency assessment in 131 ms, i.e. the same time as the original network execution time. On the other hand, the CNN instrumented with FI utilizes 23\% of the FPGA resources and performs the measurement in 632,000 ms, i.e. thousands of times (specifically, 4,824 times in this example) slower than the proposed method. This gain is composed of three components: a) processing of a single image in the CNN instrumented with APPRAISER is 0.29 ms vs 1.40 ms in the CNN instrumented with FI; b) APPRAISER pipelines the processing through the layers while FI has to break the pipeline; c) FI needs numerous iterations for each image to inject the faults (single, double or multiple) at different locations, one combination at a time, while APPRAISER uses only one iteration for each image. 

Therefore, the time difference becomes even more drastic when comparing these methods for deeper networks (determining the number of layers in the inference execution pipeline) or DNNs with a larger memory for storing weights (determining the number of potential fault locations).

\section{Conclusion}

The state-of-the-art methods for fault injection by emulation incur a spectrum of time-, design- and control-complexity problems. To overcome these issues, a novel resiliency assessment method called APPRAISER is proposed that 
applies functional approximation for a non-conventional purpose and employs approximate computing errors for its interest. 
By adopting this concept in the resiliency assessment domain, APPRAISER provides thousands of times speed-up in the assessment process, while keeping high accuracy of the analysis. In this paper, APPRAISER is validated by comparing it with state-of-the-art approaches for fault injection by emulation in FPGA. By this, the feasibility of the idea is demonstrated, and a new perspective in resiliency evaluation for DNNs is opened.


\section{Acknowledgement}
This work was supported in part by the European Union through European Social Fund in the frames of the ``Information and Communication Technologies (ICT) programme'' (``ITA-IoIT'' topic), by the Estonian Research Council grant PUT PRG1467 ``CRASHLES'' and by Estonian-French PARROT project ``EnTrustED''.
\bibliographystyle{IEEEtran}

\vspace{-0.1cm}
\bibliography{ref}

\end{document}